\documentclass[conference]{IEEEtran}
\IEEEoverridecommandlockouts
\usepackage{cite}
\usepackage{amsmath,amssymb,amsfonts}
\usepackage{algorithmic}
\usepackage{graphicx}
\usepackage{textcomp}
\usepackage{xcolor}
\usepackage{comment}
\usepackage{acronyms}
\usepackage{misc}

\usepackage[export]{adjustbox}

\usepackage[caption=false]{subfig}
\DeclareMathOperator*{\E}{\mathbb{E}}

\def\BibTeX{{\rm B\kern-.05em{\sc i\kern-.025em b}\kern-.08em
    T\kern-.1667em\lower.7ex\hbox{E}\kern-.125emX}}
\begin{document}

\title{A reinforcement learning control approach for underwater manipulation under position and torque constraints
}

\author{Ignacio Carlucho\textsuperscript{a}, Mariano De Paula\textsuperscript{b}, Corina Barbalata\textsuperscript{a}, Gerardo G. Acosta\textsuperscript{b} \\ 

icarlucho@lsu.edu, mariano.depaula@fio.unicen.edu.ar, cbarbalata@lsu.edu,  ggacosta@fio.unicen.edu.ar \\

\textsuperscript{a } \textit{Department of Mechanical Engineering,} \textit{Louisiana State University}, Baton Rouge, USA \\

\textsuperscript{b }INTELYMEC Group, Centro de Investigaciones en F\'isica e Ingenier\'ia del Centro \\ CIFICEN  \textendash \space  UNICEN \textendash \space CICpBA \textendash \space  CONICET, 7400 Olavarr\'ia, Argentina \\

}

\maketitle
\thispagestyle{plain}
\pagestyle{plain}

\begin{abstract}

In marine operations underwater manipulators play a primordial role. However, due to uncertainties in the dynamic model and disturbances caused by the environment, low-level control methods require great capabilities to adapt to change.
Furthermore, under position and torque constraints the requirements for the control system are greatly increased.
Reinforcement learning is a data driven control technique that can learn complex control policies without the need of a model. The learning capabilities of these type of agents allow for great adaptability to changes in the operative conditions. 
In this article we present a novel reinforcement learning low-level controller for the position control of an underwater manipulator under torque and position constraints. The reinforcement learning agent is based on an actor-critic architecture using sensor readings as state information.
Simulation results using the Reach Alpha 5 underwater manipulator show the advantages of the proposed control strategy.

\end{abstract}


\section{Introduction}

In the last decades the importance of underwater manipulators in marine operations has grown continuously. Most robotic underwater industrial applications are conducted with Remotely Operated Vehicles (ROV) where a human operator is tasked with the remote operation of the manipulator \cite{SIVCEV2018431}. However, due to the limited number of expert operators and the high cost of operations, the industry is migrating towards Autonomous Underwater Vehicles (AUV) \cite{YAZDANI2020103382}. In this type of scenario, a manipulator, usually electric, is mounted on the AUV and operates autonomously, however, this requires a robust and adaptable control system. Furthermore, in autonomous missions different types of operational constraints may appear, such as specific joint constraints that must be followed in order to avoid collisions \cite{Papageorgiou2011SwitchingMC}, or decreased joint torque due to faulty motors. These constraints increase the need for designing complex control systems for robust manipulation. 

One of the most used low-level controllers for manipulators is the classical Proportional Integrative Derivative (PID) controller \cite{Ziegler1993}. This is due mostly to its simplicity of use and low computational requirements. However, when it is used for controlling manipulators arms, it must cope with highly non-linear systems. This issue is aggravated in underwater environments where unknown disturbances affect the behaviour of the arm. 
Furthermore, for underwater manipulators,  controllers are used under the assumption that the arm will move slowly and as such it is possible to decouple each degree of freedom, something that is not true for every application \cite{Barbalata2018CoupledAD}.
Researchers have also turned to non-linear optimal control techniques as a viable option, since they allow to optimize a cost function under different metrics. One of these techniques is Model Predictive Control (MPC) \cite{GARCIA1989335}, used successfully for controlling a different number of underwater robots \cite{Shen2018,Bai2019ReviewAC}. However, one of the drawbacks of this technique is that it requires an accurate model of the plant in order to work properly, not a trivial matter in underwater robotics \cite{fossen1994guidance}.

Data driven control techniques have appeared as an alternative for systems with complex or unknown models. One of these techniques is Reinforcement Learning (RL) \cite{Sutton1998}. In the RL framework, the robot arm control problem can be formulated as a Markov Decision Process (MDP) \cite{Monahan1982}. 
Solving a RL problem consists in iteratively learning  a task from interactions to achieve a goal. During learning, an artificial agent (controller) interacts with the target system (arm) by taking an action (torque command), that makes the robot evolve from its current state $x_t \in \mathbb{X} \subseteq \mathbb{R}^n$ to $x_{t+1}$. The agent then receives a numerical signal $r_t$, called reward, which provides a measure of how good (or bad) the action taken at time $t$ is in terms of the observed state transition. 
Many works have used the RL paradigm for controlling AUVs in underwater environment \cite{Carlucho2018}. However, this technique has not yet been applied to underwater manipulators.

The main contribution of this work is the development of a reinforcement learning based control system for the low-level control of an electric underwater manipulator under position and torque constraints. Our reinforcement learning formulation is based on the Deep Deterministic Policy Gradient (DDPG) algorithm \cite{DDPG2016}. The proposed method uses an actor critic structure, where the actor is a function that maps system states to actions and the critic is a function that assess actions chosen by the actor. Deep Neural Networks (DNN) are used as function approximators for the actor and critic.
Results in simulation show the advantages of our proposal when controlling a simulated version of the Reach 5 Alpha manipulator, shown in Fig. \ref{fig:reach5}. The proposed controller is compared with a MPC, showing that the RL controller is able to outperform the MPC. 

The article is structured as follows, Section~\ref{sec:related_work} presents an overview of related works followed by Section~\ref{sec:dd_optimal} that introduces the basics of RL control utilized in our formulation. In Section ~\ref{sec:implementation} the details of our implementations are described, in Section~\ref{sec:results} we present the results obtained with our proposed control scheme, and finally Section~\ref{sec:conclusions} presents the overall conclusions of the proposed work.


\section{Related Works}
\label{sec:related_work}
Designing control systems under constrains considerations for robotic manipulators appeared due to the need of robots to interact with the environment. Some of the most fundamental approaches focused on designing motion/interaction control systems by using a hybrid control formulation \cite{mills1989force}, \cite{yoshikawa1987dynamic}. In this approach, the constrains are expressed based on the end-effector's working space, and are used to decide the type of control law (either a motion control law or a force regulator). Nevertheless, constraints are not imposed only by the interaction with the environment, but are also required for cases when the robotic manipulator has to adjust its working space due to obstacles in the environment, or faults in the robotic system. In \cite{zhang2019passivity} a passivity-based kinematic control law is proposed under joint velocity limits considerations. The method proposed can be adapted to different feedback control laws, but can be applied only for redundant systems. An adaptive neural-network control  for robotic  manipulators  with  parametric  uncertainties  and  motion constraints is proposed in \cite{li2016adaptive}. The simulation and experimental results with a 2 degree of freedom (DOF) planar manipulator show the velocity constraints always being respected, but steady-state errors are present. Deep neural-network approaches have become popular in the past years. An example is given in \cite{xu2019deep} where an obstacle avoidance control law is designed for redundant manipulators. The problem is reformulated as a Quadratic Programming (QP) problem in the speed level, and a deep recurrent neural network is designed to solve the QP problem in an online way. Although the simulation results show that the robot is capable of avoiding the obstacles while tracking the predefined trajectories, an experimental evaluation is not presented.

The requirements of adaptability and the difficulties with modeling have lead research towards intelligent control methods, such as reinforcement learning. Many RL methods have been previously applied to the control of manipulators \cite{Deisenroth2011LearningTC}. In \cite{Gu2017} asynchronous reinforcement learning was used to train a series of robotic manipulators to solve a door opening task. 
In \cite{Wang2020} a mobile manipulator task is solved by Proximal Policy Optimization (PPO), a well known RL algorithm \cite{Schulman2017ProximalPO}, in combination with a Deep Neural Network.  
%
Specifically for the underwater environments, RL has been used as a control techniques in several previous works \cite{CARLUCHO201871, ELFAKDI2013271, Frost2015}. However, the majority of these works focus on the of AUVs. Works utilizing RL in underwater manipulators are lacking in the literature.

\section{Reinforcement learning based control}
\label{sec:dd_optimal}


\begin{figure}[t]
    \centering
        \includegraphics[width=\columnwidth,valign=c]{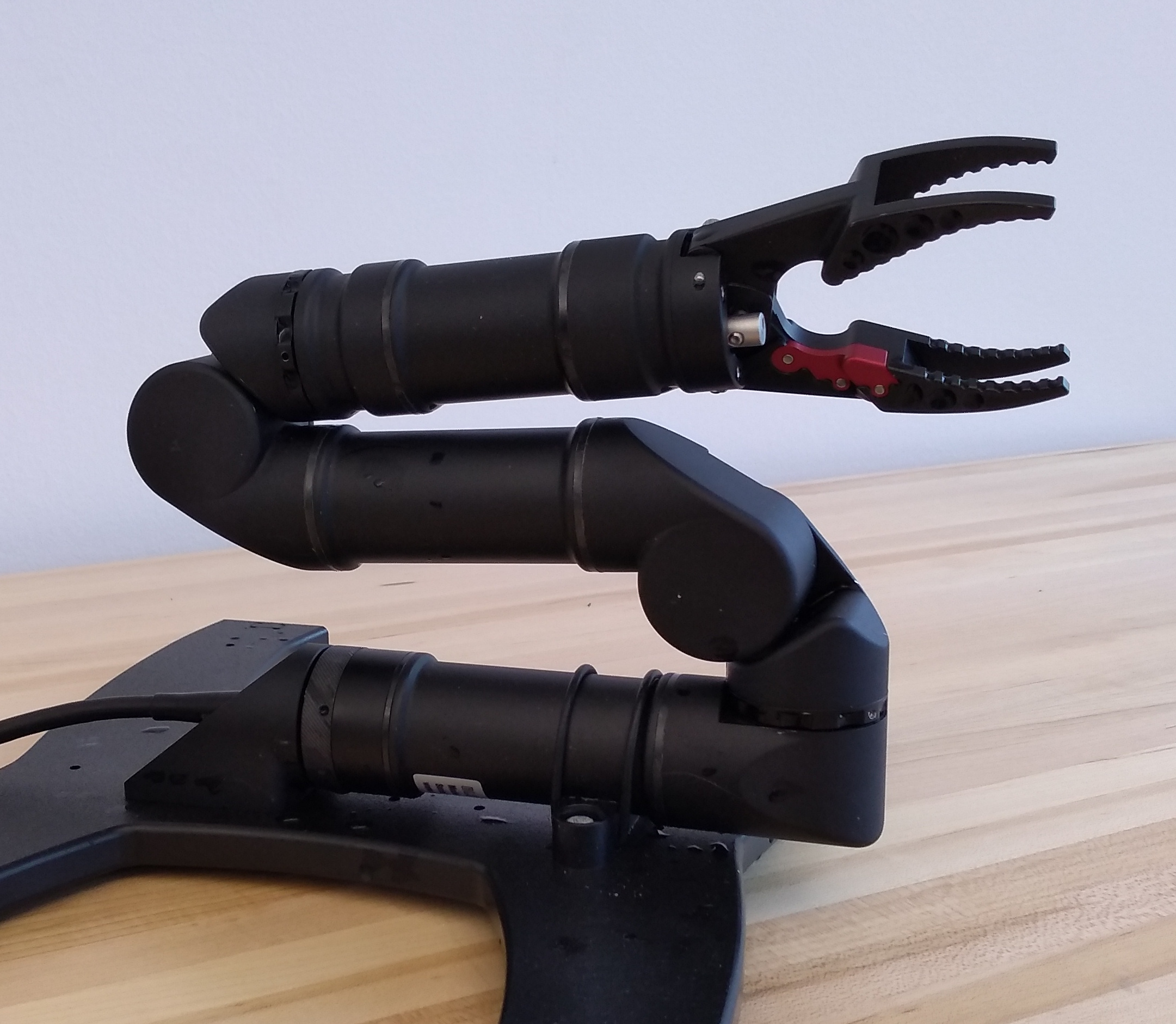}%
        \label{fig:reach5real}%
    \caption{Reach 5 Alpha underwater manipulator \cite{reachalpha}}
    \label{fig:reach5}
\end{figure}

From the point of view of classical and modern control, the design strategy of a control system is based on the complete knowledge of the dynamics of the system under study. Assuming that there is a model with the adequate capacity to describe the dynamics of the system, generally through a set of differential equations, the control problem is reduced to design a controller (or agent) capable of generating the adequate control actions for the system to achieve a given objective, goal, task or specific desired behavior. In this way, the performance capabilities of the conventional designed control systems are excessively dependent on the mathematical models used to describe the behavior of the dynamic systems to be controlled. However, underwater manipulation is a complex decision making problem in which, the presence of uncertainty in dynamics is ubiquitous and, consequently, it is of paramount importance designing and using controllers with suitable adaptation capabilities.

Markov decision processes are models for sequential decision making problems when outcomes are uncertain \cite{Monahan1982}. In our formulation, we consider a finite-horizon Markov decision process with a ${1, 2, ...,T}$ decisions and $T-1$ visited stages \cite{Hartman92}. That is, the decision (action) at time $t$ is made at the beginning of stage $t$ which corresponds to the time interval from $t$ to the next $t+1$. So, at any stage, or discrete time, $t$, the system is at a state $\mathbf{x}_t$. In this sense, we have a finite set $X$ of system states, such that $\mathbf{x}_t \, \in X, \ \forall \, t = 1, ..., T$. The decision maker observes state $\mathbf{x}_t \, \in X $ at stage $t$ and it may choose an action $\mathbf{u}_t$ from the set of finite allowable actions $U$ generating cost $L(\mathbf{x}_t,\mathbf{u}_t)$. Moreover, we let $p(\cdot|\mathbf{x}_t,\mathbf{u}_t)$ denote the probability distribution or transition probabilities of obtaining states $\mathbf{x}'=\mathbf{x}_{t+1}$ at stage $t+1$.

A deterministic Markovian decision rule at state $\mathbf{x}_t$ is a function $\psi_t:\, \mathbf{x}_t \to \mathbf{x}_t$ which maps the action choice given at state $\mathbf{x}_t$. It is called deterministic because it chooses an action with certainty and Markovian (memoryless) since it depends only on the current system state. We let $D_t$ denote the set of possible deterministic Markovian decision rules at stage $t$. $D_t$ is a subset of more general rules where the action may depend on the past history of the system and actions may not be chosen with certainty but rather according to a probability distribution. 

A policy or strategy specifies the decision rules to be used at all stages and provides the decision maker with a plan of which action to take given stage and state. That is, a policy $\pi$ is a sequence of decision rules and we restrict ourselves to ranking policies $\pi_t$ belonging to the set $D_t$ of deterministic Markov policies (if randomized policies were included, the set of policies would not be countable). In some problems, the decision maker only focuses on this subset of policies, e.g. because randomized policies are hard to manage in practice or restrictions in management strategy. Moreover, if the states at a given time step $t$ corresponds to different physical locations implementation of policies having a single action at each location may only be acceptable.


Under the above summarized framework of Markov decision processes, the reinforcement learning can be formalized where an RL agent located in its environment chooses an action, from the set of available ones, at every discrete time step $t$ based on the current state of the system $\mathbf{x}_t$. In return, at each time step, the agent receives a reward signal that quantifies the quality of the action taken in term of the goal of the control task. In this paper we only consider one criterion of optimality, namely the expected total cost criterion, so the objective is to obtain an optimal policy $\pi^*$ that satisfies: 
\begin{equation}
  L^* = \max L_\pi = \max \E_\pi \{ R_t | \mathbf{x}_t =  \mathbf{x} \} 
\end{equation}


\noindent where $r_t$ is the instantaneous reward obtained at time step $t$ and $R_t$ is the cumulative reward, such that ${R_t = \sum_{k=0}^{\infty} \gamma^k r_{t+k+1}}$. 

The basic RL algorithms discussed in the literature have been extensively developed to solve reinforcement learning problems without need of a dynamic model and when the state spaces and action are finite, which means that the value functions support a tabular representation \cite{Sutton1998,CARLUCHO2017,CARLUCHO2019}. In order to find an approximate solution to the control problem it is possible to obtain a discretized form of the space of states and/or actions \cite{LOVEJOY1991,CRESPOSUN2000,CRESPOSUN2003} and then applying the RL algorithms that use discrete spaces. However, as the granularity of the representation increases, the computational implementations suffer the so-called curse of dimensionality, which consists of an exponential increase in the computational complexity of the problem due to the increase in the dimension (or number) of the state-action pairs to be selected. This makes it impossible to construct a value function for the problem in question, since the agent has a low probability of "visiting" the same state, or a state-action pair, more than once depending on whether it is working with the state value function or value-action function, respectively.

In the underwater manipulation problem we have to deal with dynamic systems where the states and the applied actions are defined in real domains (continuous spaces) which imposes an important limitation for tabular representation of the value functions. To overcome this drawback, functional approximation techniques have emerged including inductive models to attempt generalizing the value function. Since a few years ago powerful brain inspired deep neural networks \cite{LeCun2015} have been introduced as functions approximations into the RL framework giving rise to deep reinforcement learning methodologies \cite{Mnih2015}. 
For instance, the Deep Deterministic Policy Gradient (DDPG) algorithm \cite{DDPG2016} is one of the most spread deep RL algorithms that utilizes an actor-critic formulation together with neural network as function approximators to obtain a deterministic optimal policy.
In the actor-critic formulation, the role of the actor is to select an action based on the policy, such that $\mathbf{u} = \pi(\mathbf{x}_t)$. The critic on the other hand, gives feedback of how good or bad the selected action was.  
In the DDPG algorithm the state-action value function $Q(\mathbf{x}_t, \mathbf{u}_t)$ is used as a critic. This function is defined as: 
\begin{equation}
Q^\pi(\mathbf{x}_t, \mathbf{u}_t) = \E \{ R_{t} | \mathbf{x}_t, \mathbf{u}_t\} = \E \{ \sum_{k=0}^{\infty} \gamma^k r_{k+t+1} | \mathbf{x}_t, \mathbf \}
\end{equation}

\noindent The update to the state-action value function can then be performed as: 
\begin{equation}
Q^w(\mathbf{x}_t, \mathbf{u}_t) = \E \{ r_{x_t,u_t} + \gamma Q^w(\mathbf{x}_{t+1}, \mathbf{u}_{t+1}) \}
\end{equation}

\noindent where \(Q^w\) is a differentiable parameterized function, so that $Q^w \approx Q^{\pi}$.

For the actor we consider a function $\pi$ that parameterizes states directly into actions with parameters \(\theta\), thus $ \pi(x_t|\theta)$. And we define a performance objective function ${L(\pi_{\theta})= \E \{r^{\gamma}|\mu \}}$ and a probability distribution $\rho$, then the performance as an expectation can be written as: 

\begin{equation}
L(\mu_{\theta}) = \int \rho^{\mu}r(\mathbf{x}_t,\mu)dx = \E [ r(\mathbf{x}_t,\mu_{\theta}(\mathbf{x}_t))]
\end{equation}
and by applying the chain rule to the expected return, we can then write: 

\begin{equation} \label{eq:ddpg}
\centering
\nabla_{\theta} L = \E [\nabla_{\theta}\mu_{\theta}(\mathbf{x}) \nabla_{\mathbf{u}} Q^{\mu}(\mathbf{x},\mathbf{u})]
\end{equation}

\noindent where Eq. \eqref{eq:ddpg} is the deterministic gradient of policies, as demonstrated in \cite{Silver2014}.
As indicated previously, both the actor and the critic can be represented by function approximators, where deep neural networks are commonly used since they allow to work with continuous state spaces.
However, the nonlinearities of these networks and the training procedures used made it difficult for algorithms to converge, however, recent events have repaired these problems. The main causes of the lack of convergence were the correlation of the samples used for training and the correlation between the updates of the network $ Q $ \cite{Fujimoto2018}.

The first of these issues was addressed by implementing a replay buffer that stored state transitions, the actions applied, and the rewards earned. The agent is then trained using mini-batches of transitions that are randomly selected from the replay buffer \cite{Ioffe2015}. The second problem was solved by incorporating target networks, which are a direct copy of the actor's and critic's networks called $ \pi'$ and $Q' $, with the parameters $ \theta'$ and $ \omega'$ respectively, and which are periodically updated according to the parameter $ \tau$, so that $ \theta' \gets \theta \tau + (1- \tau) \theta'$ and $ \omega' \gets \omega \tau + (1- \tau) \omega'$ where $ \tau << 1 $.




\section{Implementation details}
\label{sec:implementation}
\begin{figure*}[t]
    \centering
    \subfloat[Joint Position]{%
        \includegraphics[height=4.5cm,valign=c]{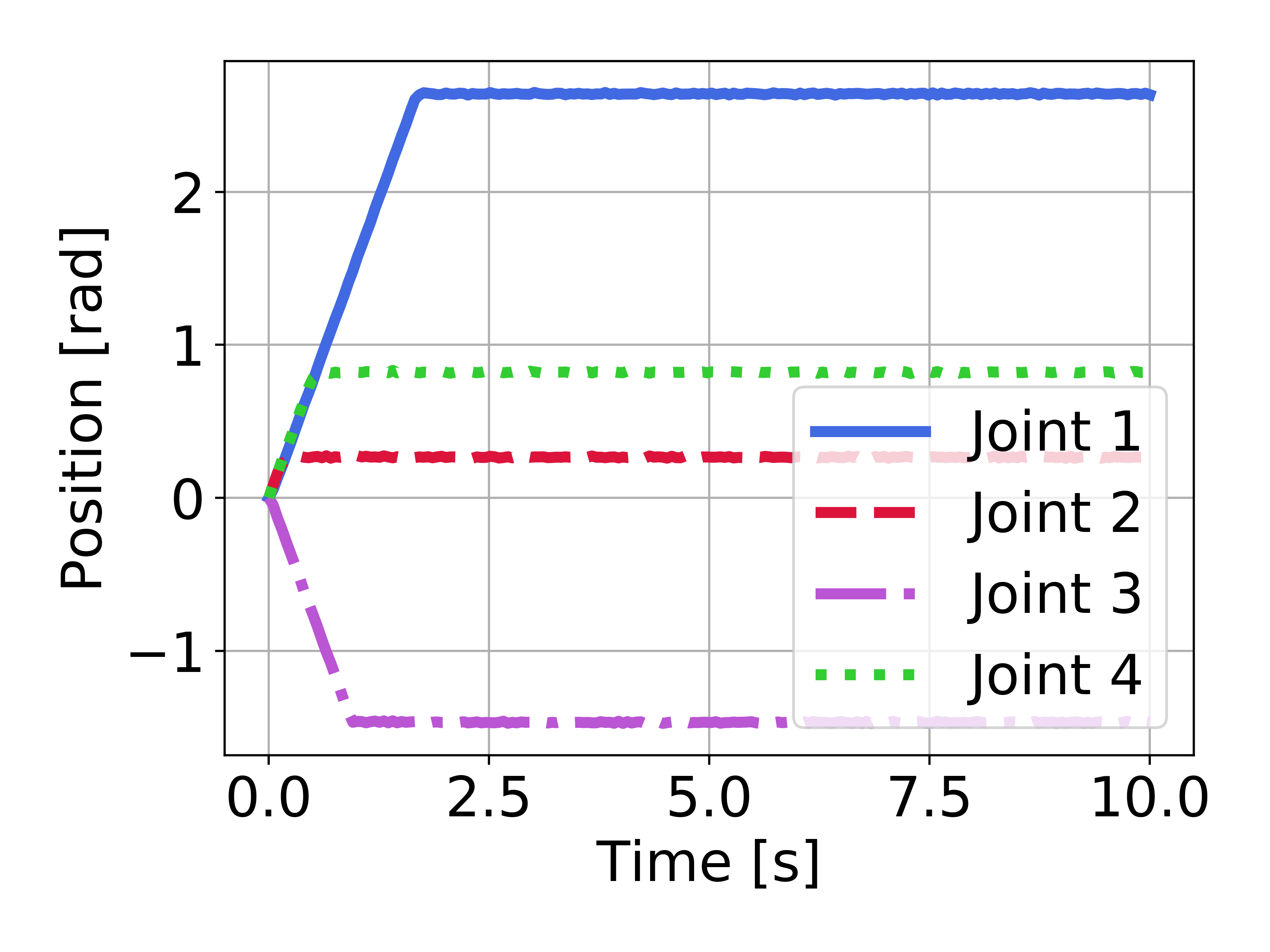}%
        \label{fig:result1a}%
          } \hfil
    \subfloat[Torque Output]{%
        \includegraphics[height=4.5cm,valign=c]{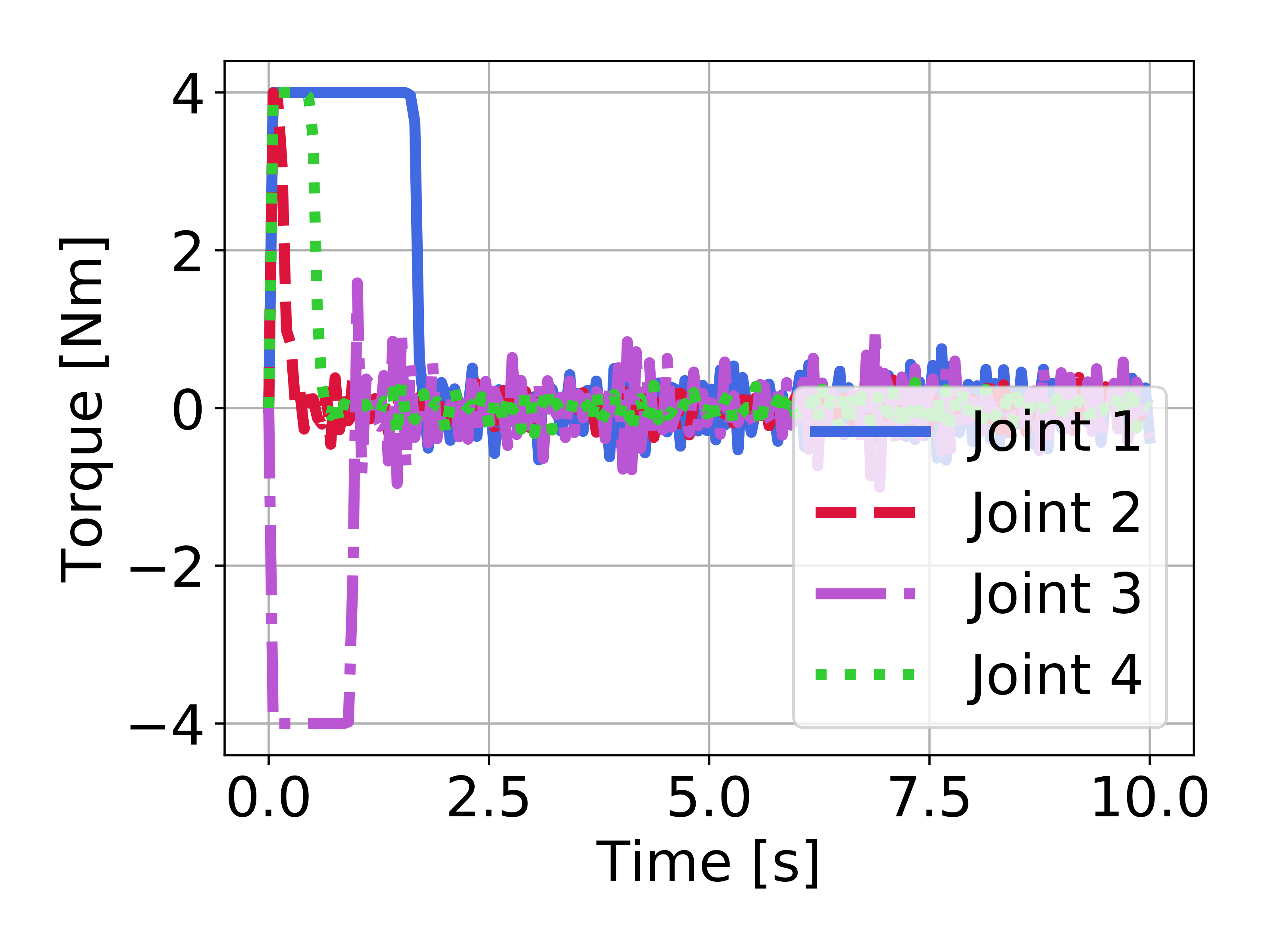}%
        \label{fig:result1b}}%
    \subfloat[Joint Errors]{%
        \includegraphics[height=4.5cm,valign=c]{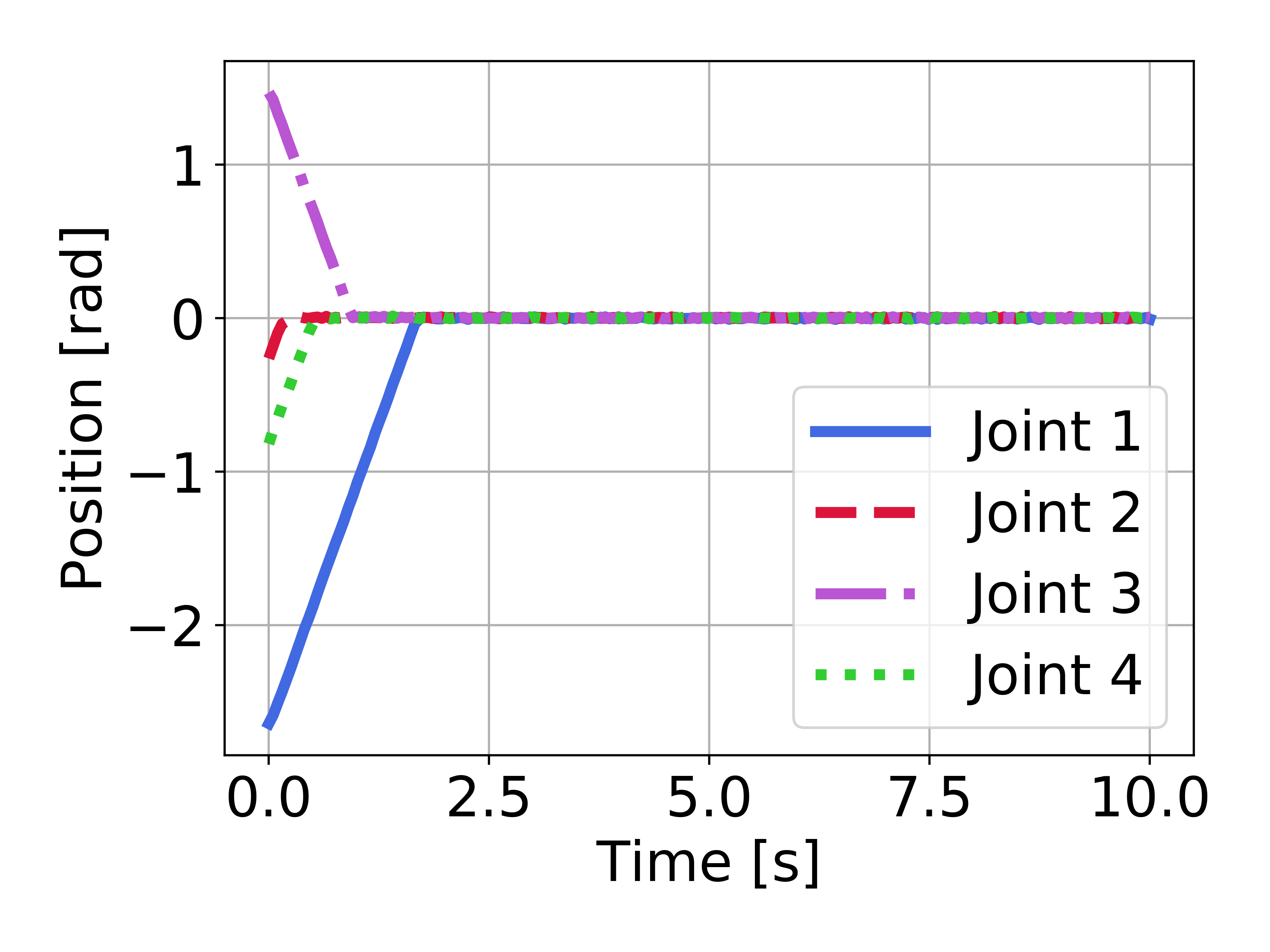}%
        \label{fig:result1c}}%
    \caption{Test 1 of the proposed RL algorithm with $\mathbf{x}_{ref} = (2.64, 0.26, -1.47, 0.82)$ [rad]}
    \label{fig:result1}
\end{figure*}

\begin{figure*}[t]
    \centering
    \subfloat[Joint Position]{%
        \includegraphics[height=4.5cm,valign=c]{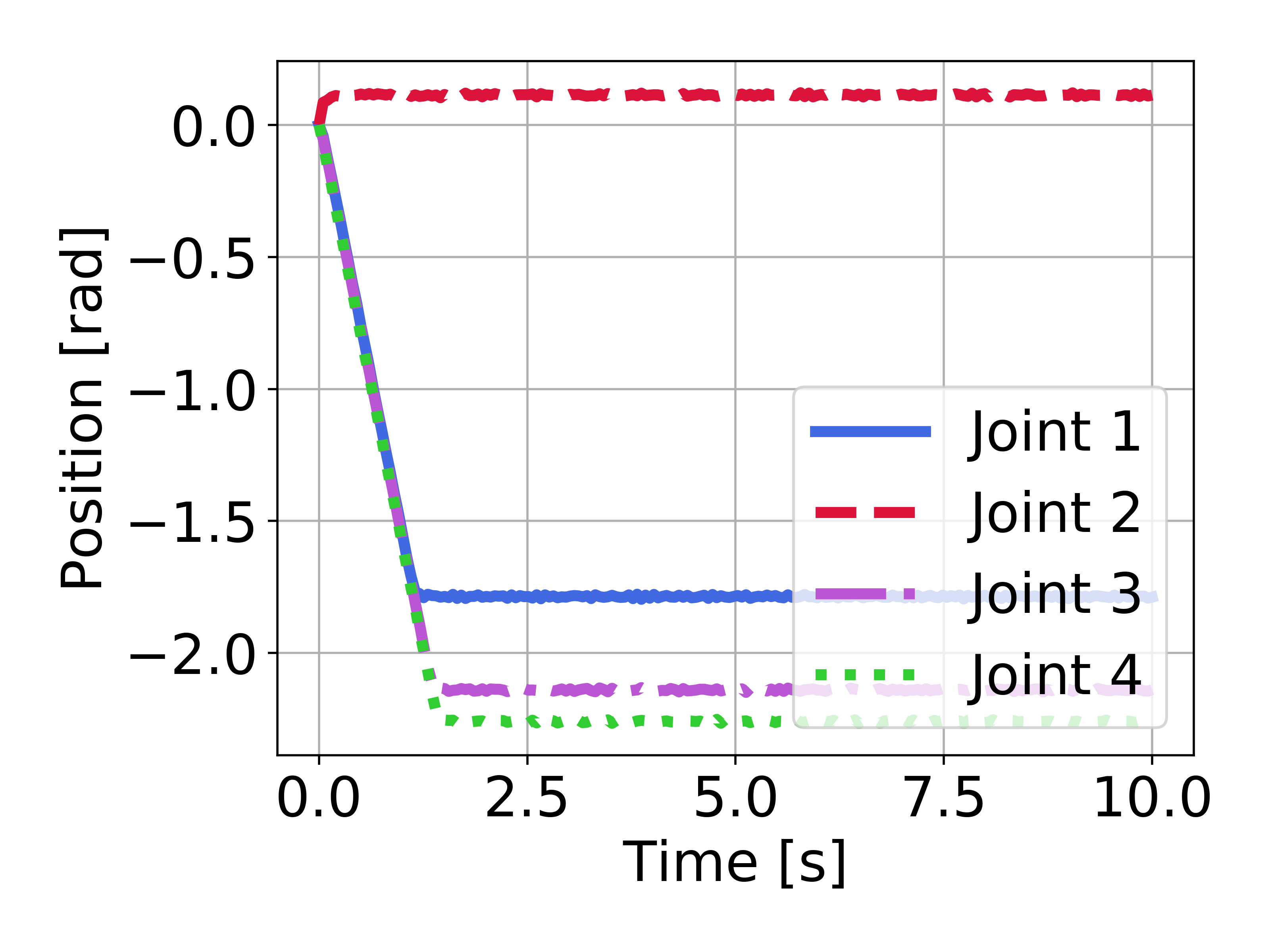}%
        \label{fig:result2a}%
          } \hfil
    \subfloat[Torque Output]{%
        \includegraphics[height=4.5cm,valign=c]{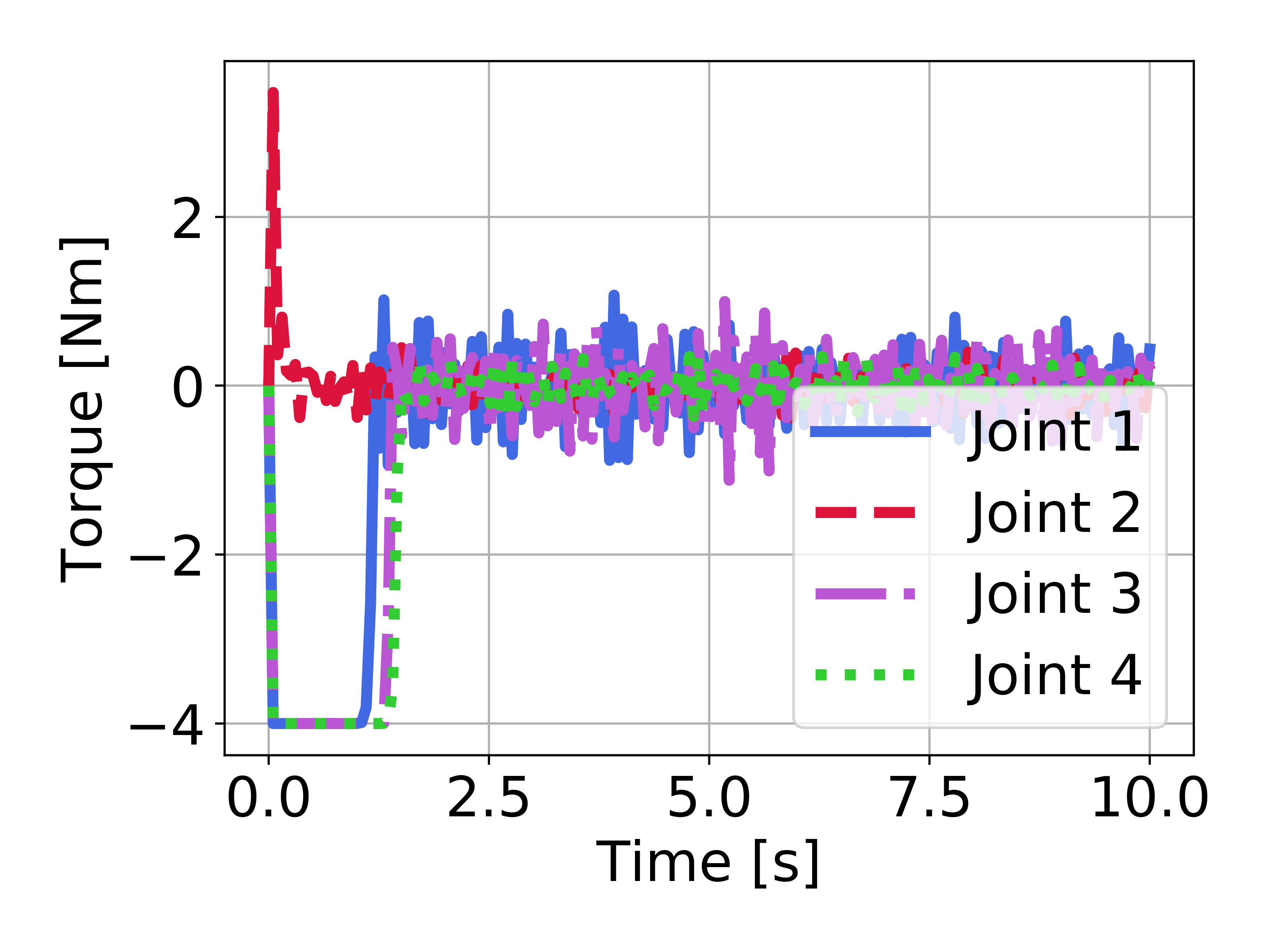}%
        \label{fig:result2b}}%
    \subfloat[Joint Errors]{%
        \includegraphics[height=4.5cm,valign=c]{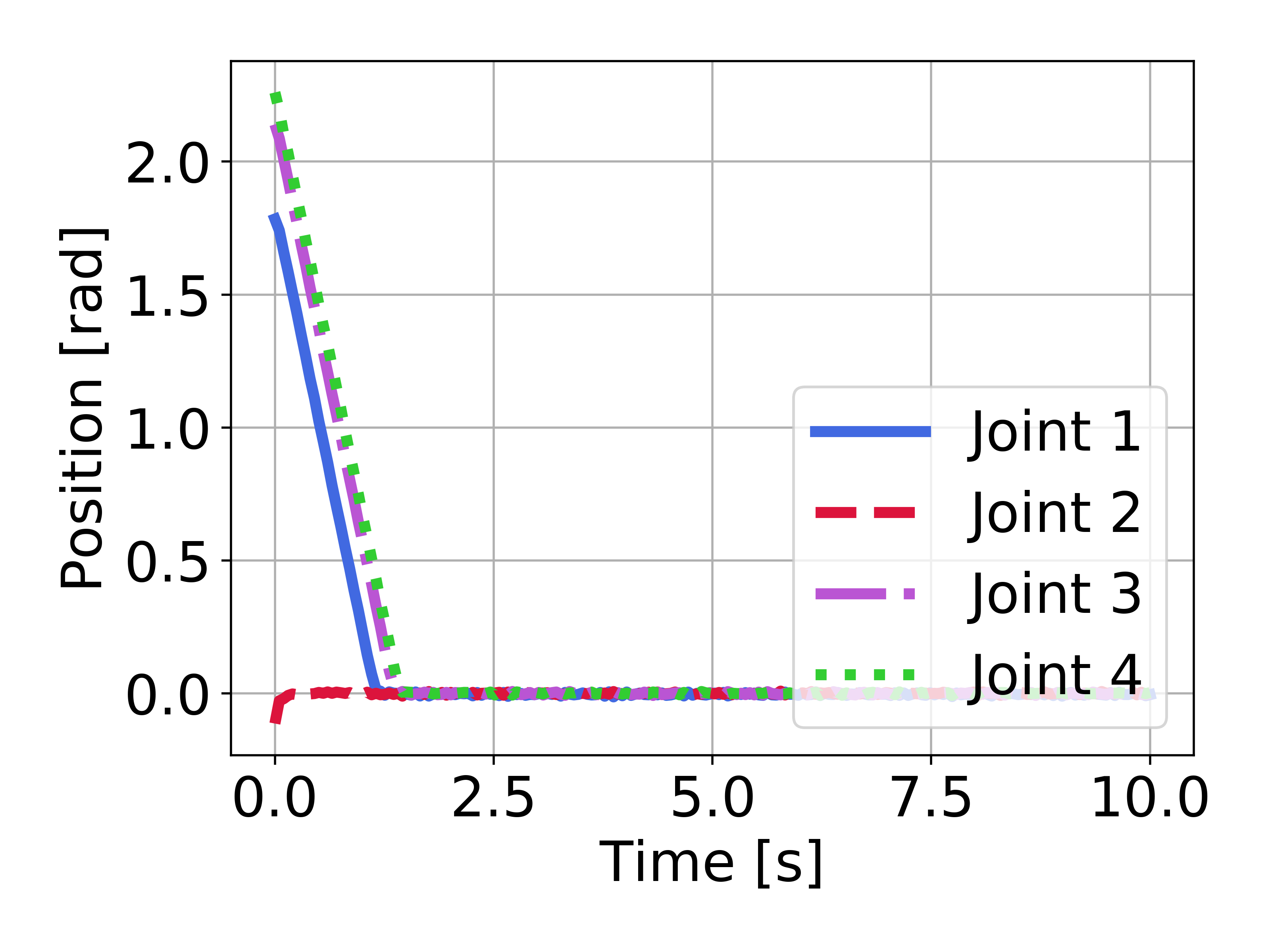}%
        \label{fig:result2c}}%
    \caption{Test 2 of the proposed RL algorithm with $\mathbf{x}_{ref} = (-1.78, 0.11, -2.14, -2.26)$ [rad]}
    \label{fig:result2}
\end{figure*}

\begin{figure*}[t]
    \centering
    \subfloat[Joint Position]{%
        \includegraphics[height=4.5cm,valign=c]{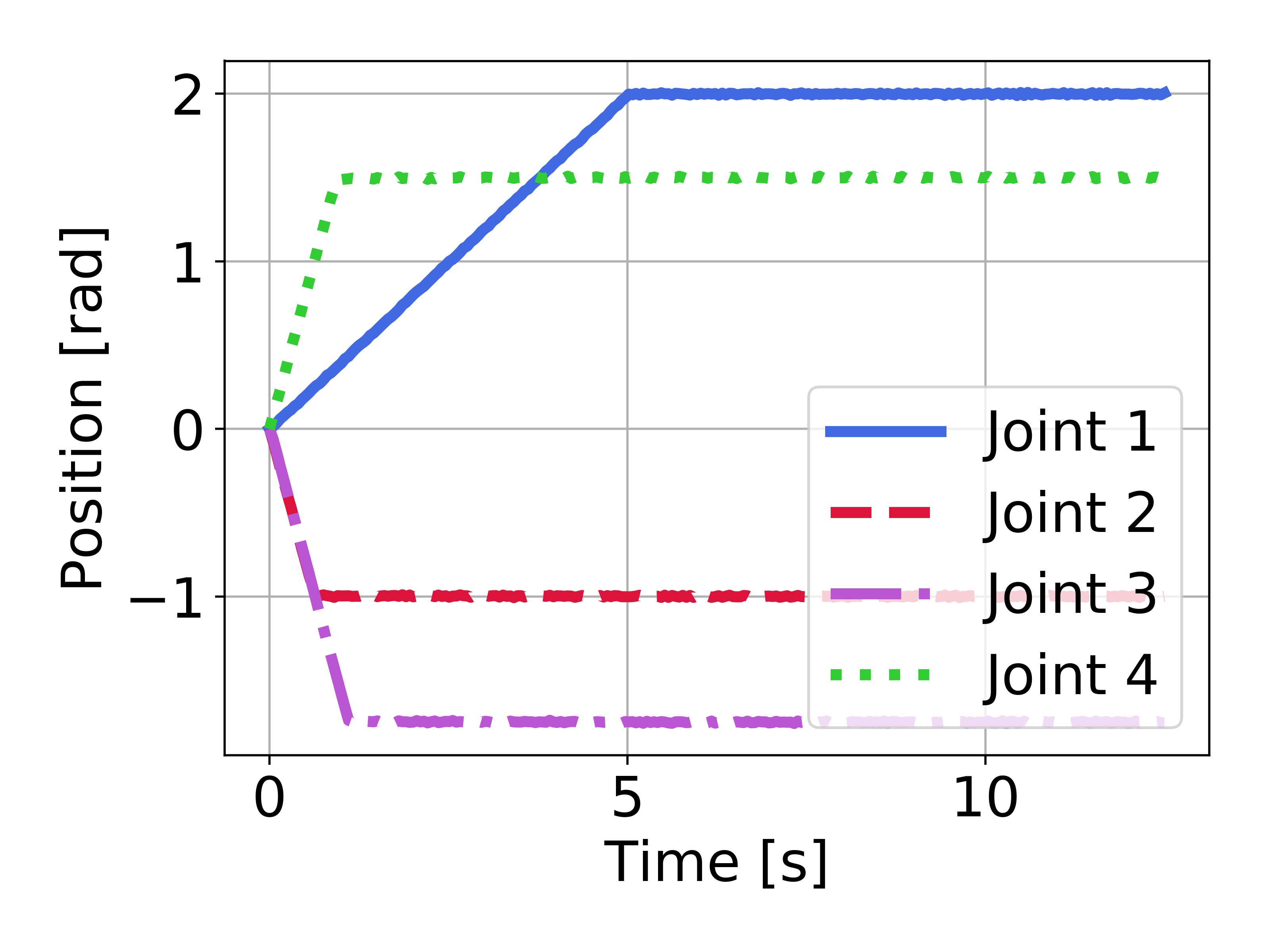}%
        \label{fig:resultta}%
          } \hfil
    \subfloat[Torque Output]{%
        \includegraphics[height=4.5cm,valign=c]{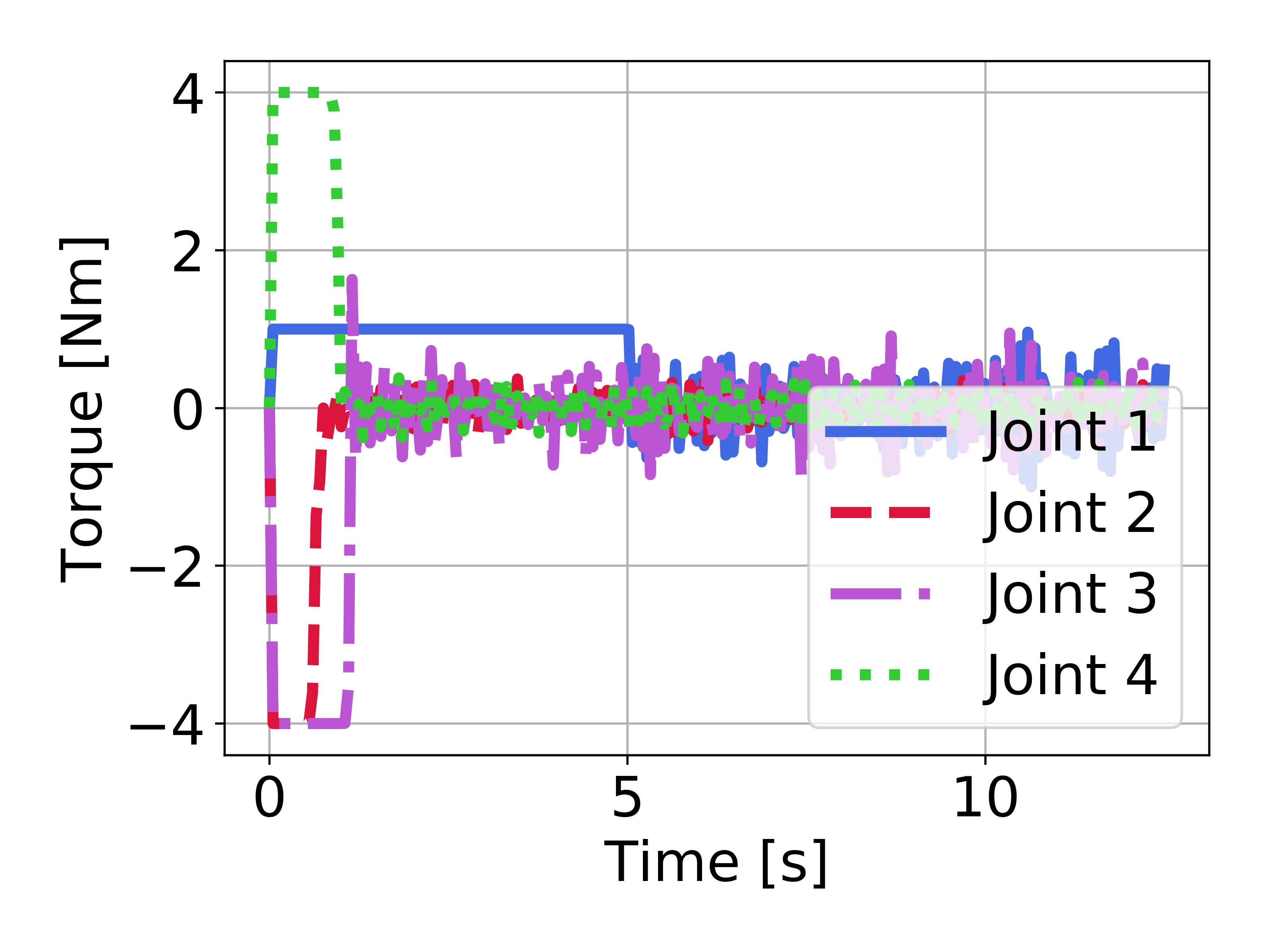}%
        \label{fig:resulttb}}%
    \subfloat[Joint Errors]{%
        \includegraphics[height=4.5cm,valign=c]{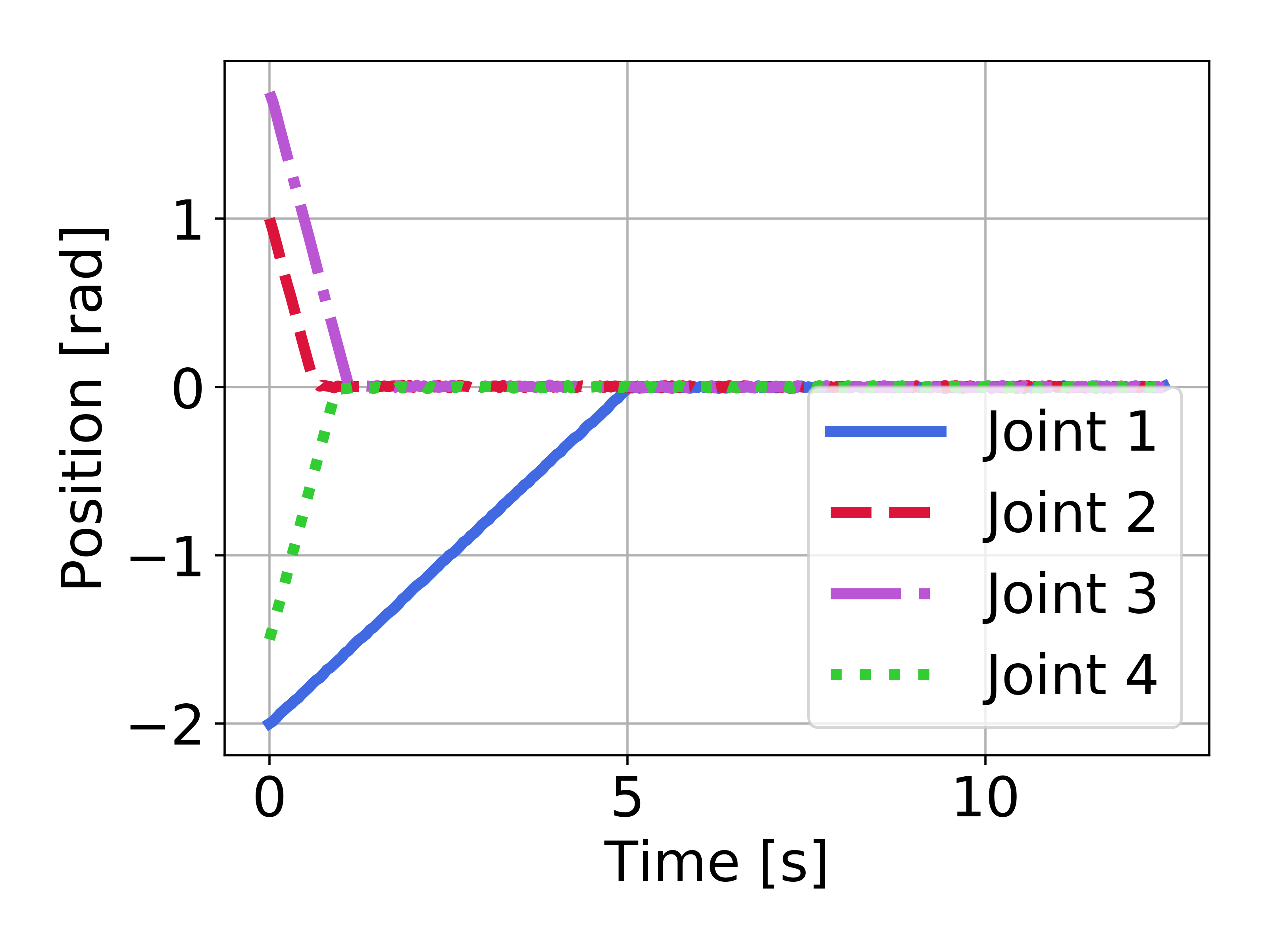}%
        \label{fig:resulttc}}%
    \caption{Test 3 of the proposed RL algorithm under Torque constraints with $\mathbf{x}_{ref} = (2.,-1., -1.75, 1.5)$ [rad]}
    \label{fig:resultt}
\end{figure*}

A simulated version of the Reach Alpha 5 manipulator is used. The Reach Alpha 5 is a 5DOF underwater manipulator, capable of lifting 2kg, and is able to operate in depths up to 300m. The manipulator is shown in Fig. \ref{fig:reach5}.

For our proposed formulation, the state ($\mathbf{x}_t$) of the RL agent is determined by the joint position ($\mathbf{q} \in \mathbb{R}^n$) in [rads] and joint velocity $\mathbf{\dot{q}} \in \mathbb{R}^n$ in [rads/s], together with the desired joint position ($\mathbf{q}_{req}  \in \mathbb{R}^n $) in [rad], such that the state is determined as:  $\mathbf{x}_t = [\mathbf{q}_t, \mathbf{\dot{q}}, \mathbf{q}_{req}]$, with $n$ being the DOFs of the manipulator. The goal of the agent is to achieve a determined joint position, where the request comes to the  agent by higher a layer in the control hierarchy.   

A fully connected feed forward network is used for both the actor and critic with two hidden layers of 400 and 300 units each. As activation function Leaky ReLus are used for the hidden networks, with Tanh used for the output neurons of the actor. The learning rate used is $0.0001$ and $0.001$ for the actor and critic respectively, with Adam used as an optimizer. A decay rate of $0.96$ is applied to each learning rate after 100 thousands training steps.

In order to be able to achieve the required position and torque constraints, the reward function was developed as follows: If the position is within the allowed bounds the reward is a Gaussian function that penalizes the agent when the joint position is not close to the request and gives a positive reward when the position matches or is close to the request. On the other hand, when the agent goes over the allowed bounds it is penalized with a high negative number ($-10$). Formally the reward is defined as follows: 

\begin{align} \label{eq:reward}
    r_t = 
\begin{cases}
  -1 + e^{ - \frac{1}{2}(\frac{\mathbf{x}- \mathbf{x}_{ref}}{\sigma})^2},   & \text{if } \mathbf{x}_{min} < \mathbf{x}_t < \mathbf{x}_{max} \\
  -10,  & \text{otherwise}
\end{cases}
\end{align}
\noindent with $\sigma$ being a parameter that shapes the Gaussian function. For the experiments presented here we utilized  $\sigma = 0.018$.

For the training of the agent, a different random goal was selected in the allowed work space for each epoch of training. The agent was trained for a total of 2000 epochs, with each epoch lasting 20 seconds of real time operation. Initially random noise is added to the actions of the agent to allow for exploration, such that $\mathbf{u}_t = \pi(\mathbf{x}_t) + \epsilon N $, with $N$ being Ornstein–Uhlenbeck noise and $\epsilon$ linearly decaying over time from $1$ to $0.1$. 
The minibatch size used for training is 64, with $\tau = 0.001$ and $\gamma = 0.99$.

\section{Results}
\label{sec:results}
For the presented results the agent was trained as previously stated and the policy is now executed without the addition of exploratory noise, effectively making $\epsilon = 0$. Furthermore, the nonlinear model has been degraded with some parameters changed randomly to test the adaptability of the agent and random noise is introduced to the velocity and position readings. While the Reach 5 Alpha has 5 DOF, the last degree of freedom corresponds to the gripper joint, which we are not interested in controlling, as such all results are shown for the first 4 DOF.

A simulation using the trained RL agent was ran on the Reach Alpha manipulator under normal operative conditions with a reference joint position of ${\mathbf{x}_{ref} = (2.64, 0.26, -1.47, 0.82)}$ [rad]. Fig. \ref{fig:result1a} shows the joint position while being controlled by the RL agent, Fig. \ref{fig:result1b} shows the control actions and Fig. \ref{fig:result1c} shows the position errors. In Fig. \ref{fig:result1a} it can be seen how the agent reaches the desired position in less than two seconds, without any overshoot, even when the requested position requires a long rotation of over two radians in Joint 1. The lack of overshoots demostrates how the agent is capable of behaving without breaking any of the position constrains imposed during training. The agent utilizes the maximum torque initially available, as can be seen in Fig. \ref{fig:result1b}, and then utilizes small corrections to keep the joints in position. Fig. \ref{fig:result1c} shows how the errors are rapidly reduced, with no steady state error present.

Another example shows the behaviour of the arm when a completely different reference point is selected, ${\mathbf{x}_{ref} = (-1.78, 0.11, -2.14, -2.26)}$ [rad]. Fig. \ref{fig:result2a} shows the obtained position when using the RL controller. As can be seen the requested position is reached again in under two seconds, without any overshoot. Again the agent utilizes high levels of torque initially, with lower levels after the requested joints position have been reached, as depicted in \ref{fig:result2b}. Furthermore, no steady state error is present as ilustrated in \ref{fig:result2c}.

The example presented here aims to test the behavior of the agent under torque constraints. In this example, the torque output of Joint 1 is reduced by 75\%, with a desired requested position of $\mathbf{x}_{ref} = (2.,-1., -1.75, 1.5)$ [rad]. In Fig. \ref{fig:resultta} the achieved position are shown, where it can be seen that the agent is capable of rapidly reaching the desired positions for Joints 2, 3 and 4, while Joint 1 takes around 5 seconds due to the new restrictions in torque, but no overshoot is present. This can also be seen in Fig. \ref{fig:resulttc} where the errors are shown, where Joint 1 takes longer to reach the request, however it can also be seen that no steady state error or overshoot are present. Additionally, Fig. \ref{fig:resulttb} shows the torque output, where the reduced torque of Joint 1 can be clearly seen.

\begin{figure*}[t]
    \centering
    \subfloat[RL: Joint Position]{%
        \includegraphics[height=4.5cm,valign=c]{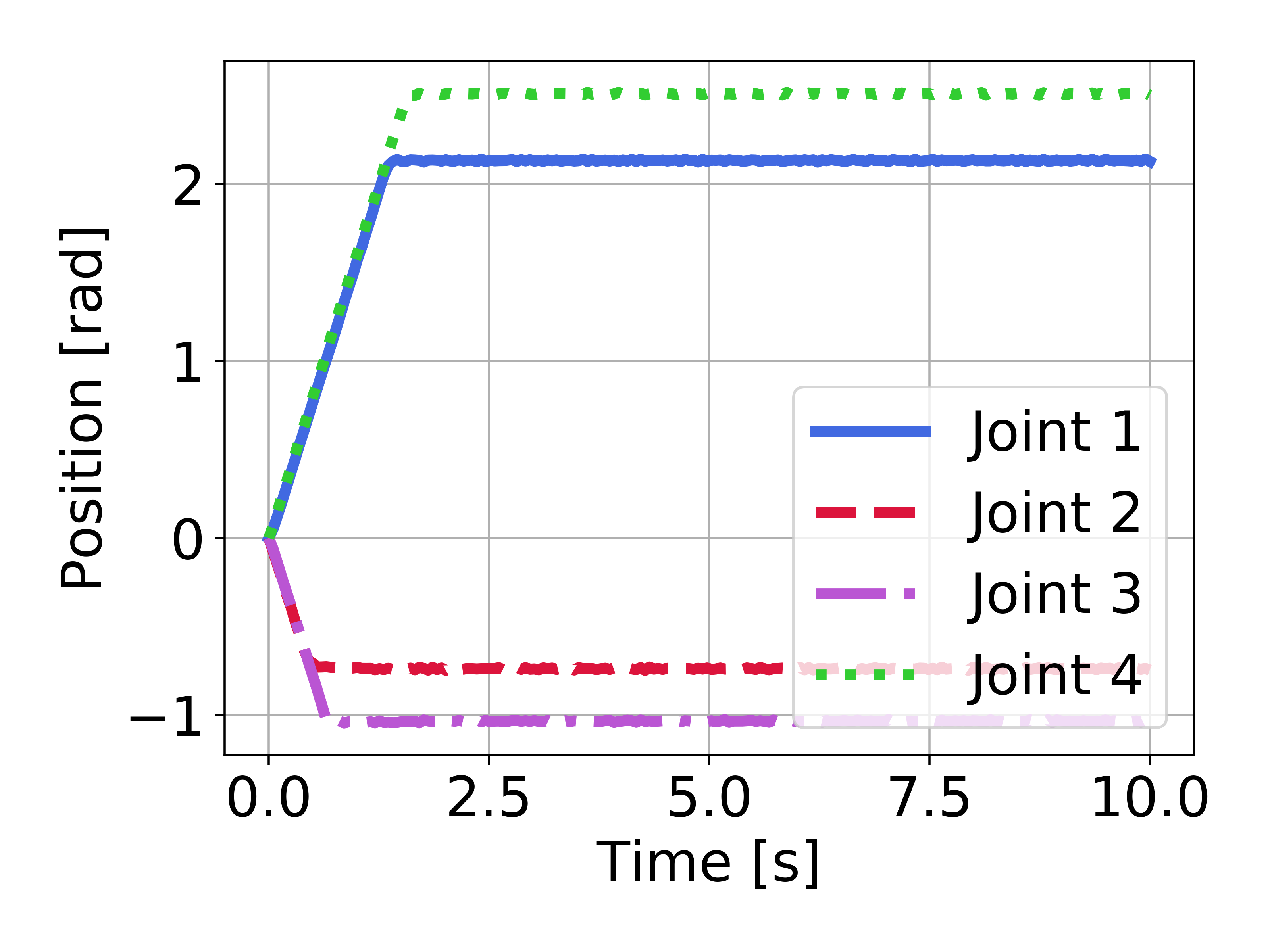}%
        \label{fig:result_compare_rl_pos}%
          } \hfil
    \subfloat[RL: Torque Output]{%
        \includegraphics[height=4.5cm,valign=c]{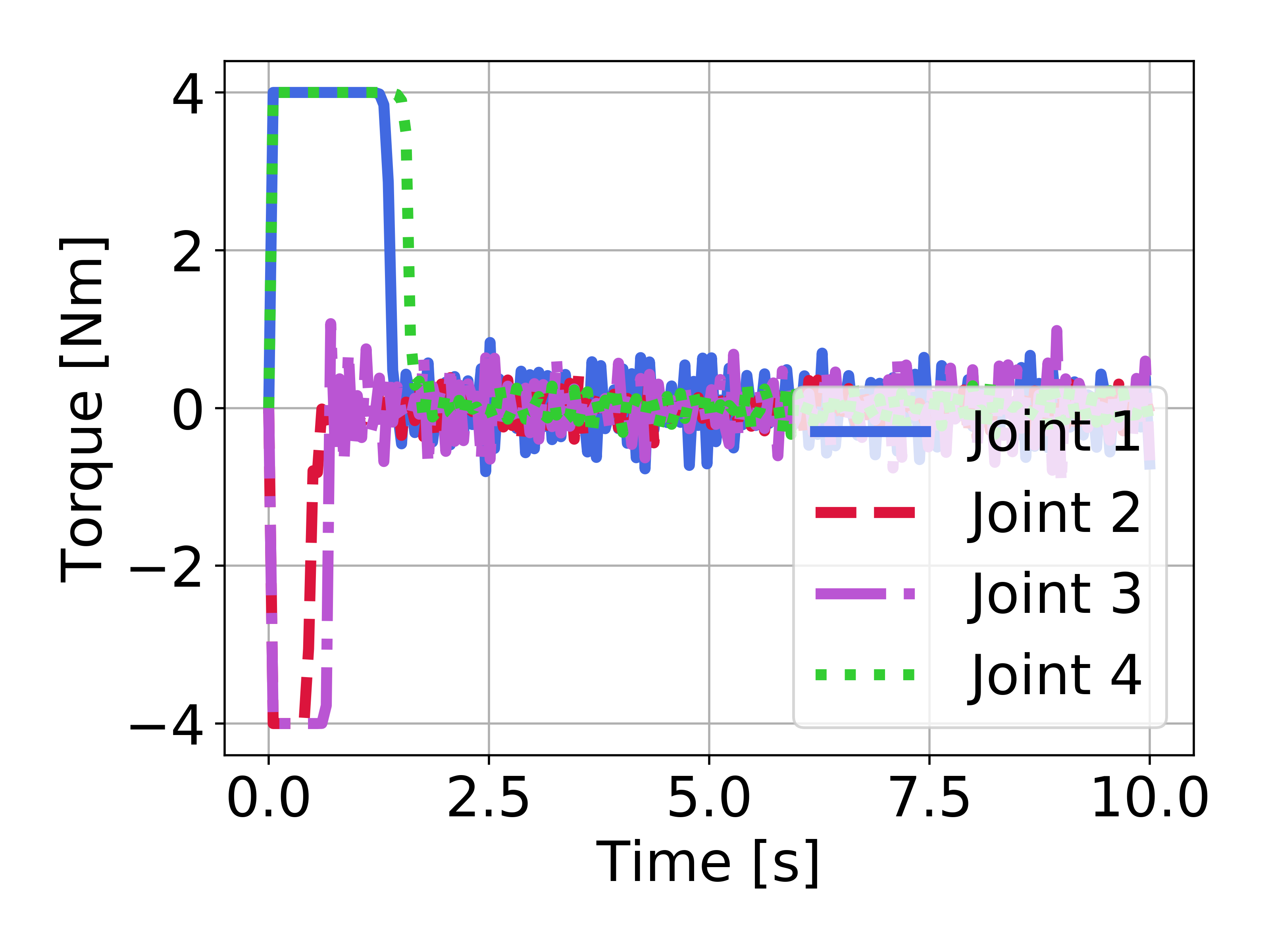}%
        \label{fig:result_compare_rl_act}}%
    \subfloat[RL: Joint Errors]{%
        \includegraphics[height=4.5cm,valign=c]{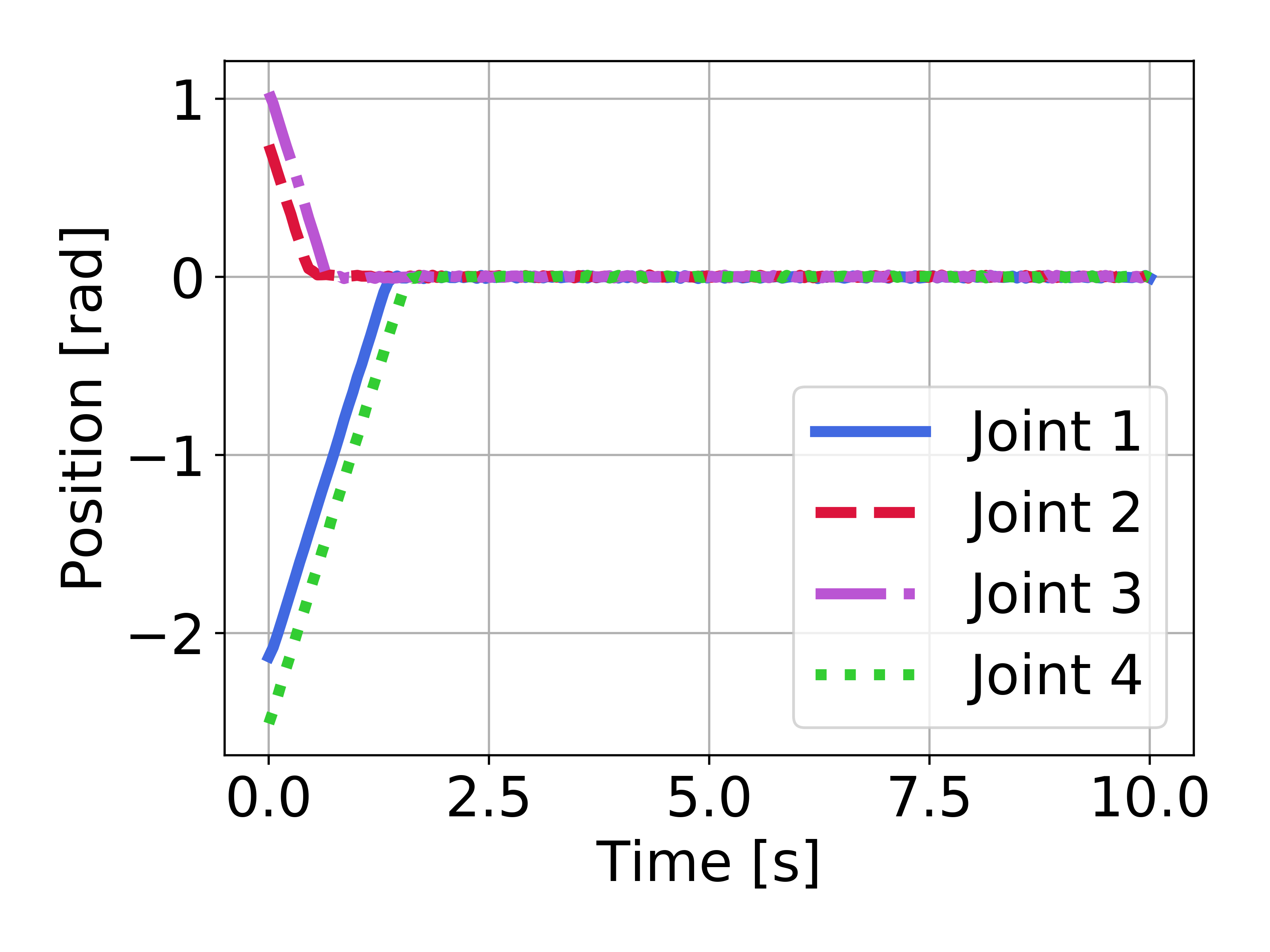}%
        \label{fig:result_compare_rl_err}}%
    \caption{Comparative results using RL for $\mathbf{x}_{ref} = (2.13, -0.74, -1.03,  2.51)$ [rad]}
    \label{fig:result_compare_rl}
\end{figure*}

\begin{figure*}[t]
    \centering
    \subfloat[MPC: Joint Position]{%
        \includegraphics[height=4.5cm,valign=c]{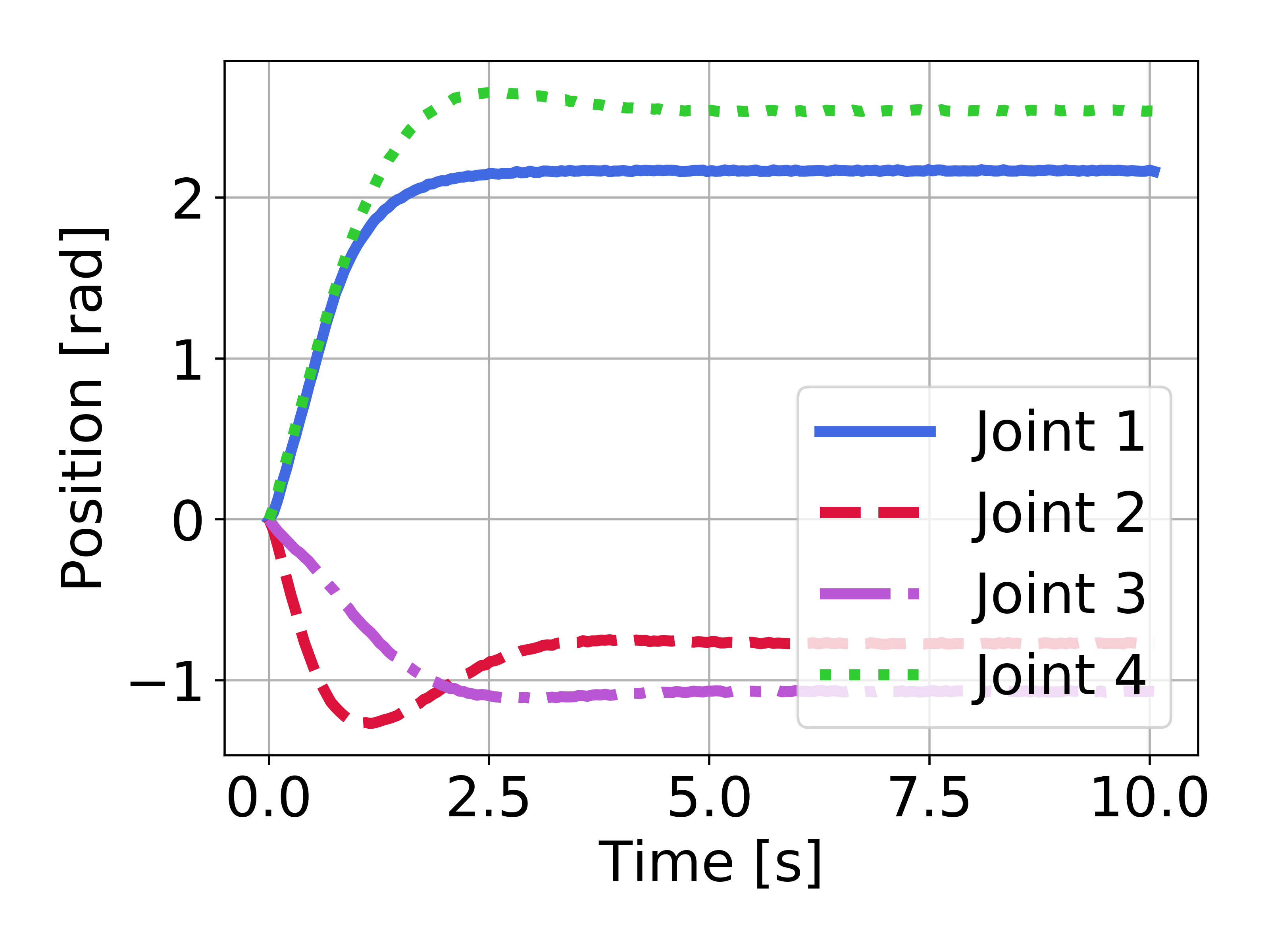}%
        \label{fig:result_compare_mpc_pos}%
          } \hfil
    \subfloat[MPC: Torque Output]{%
        \includegraphics[height=4.5cm,valign=c]{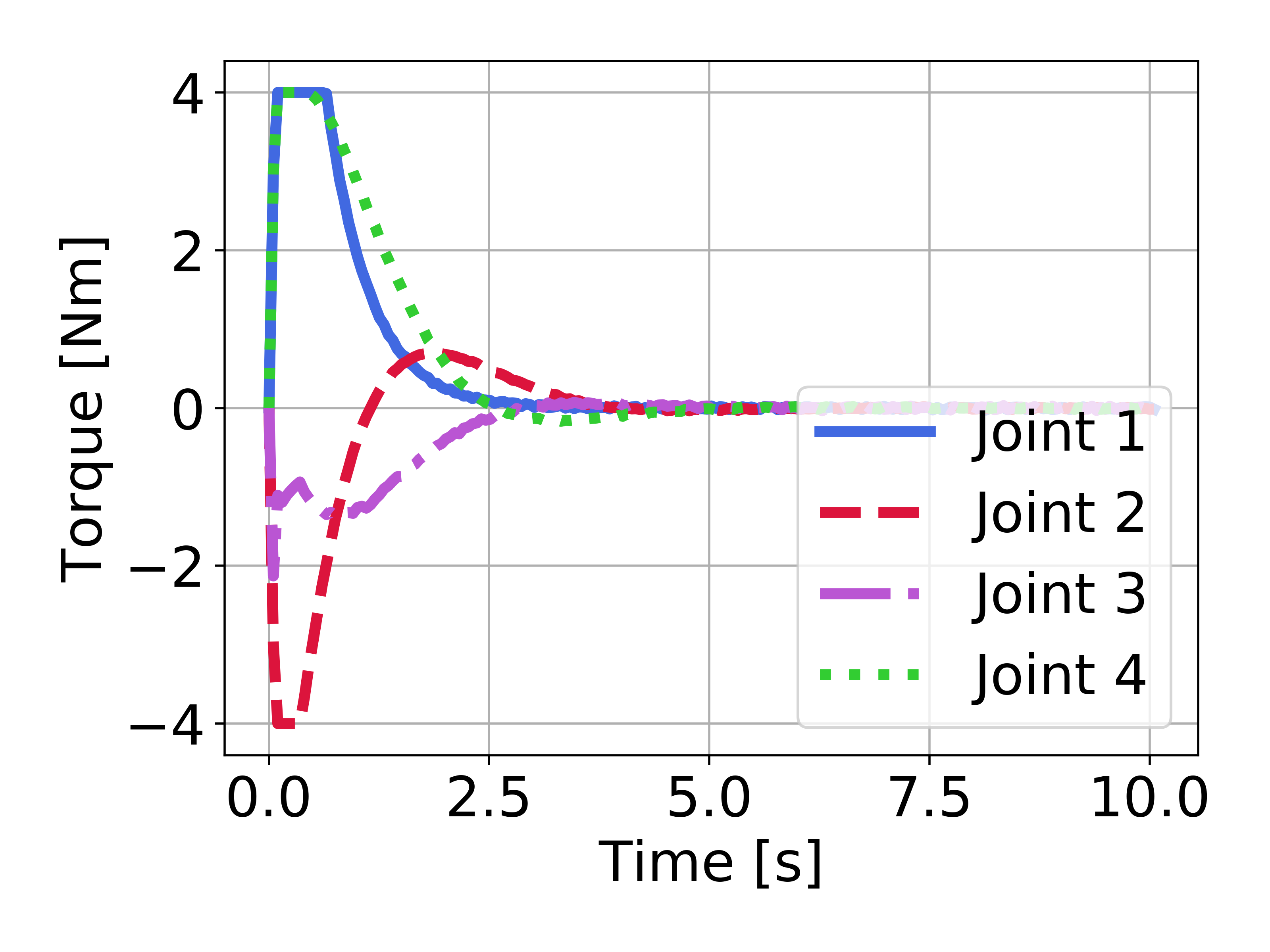}%
        \label{fig:result_compare_mpc_act}}%
         \subfloat[MPC: Joint Errors]{%
        \includegraphics[height=4.5cm,valign=c]{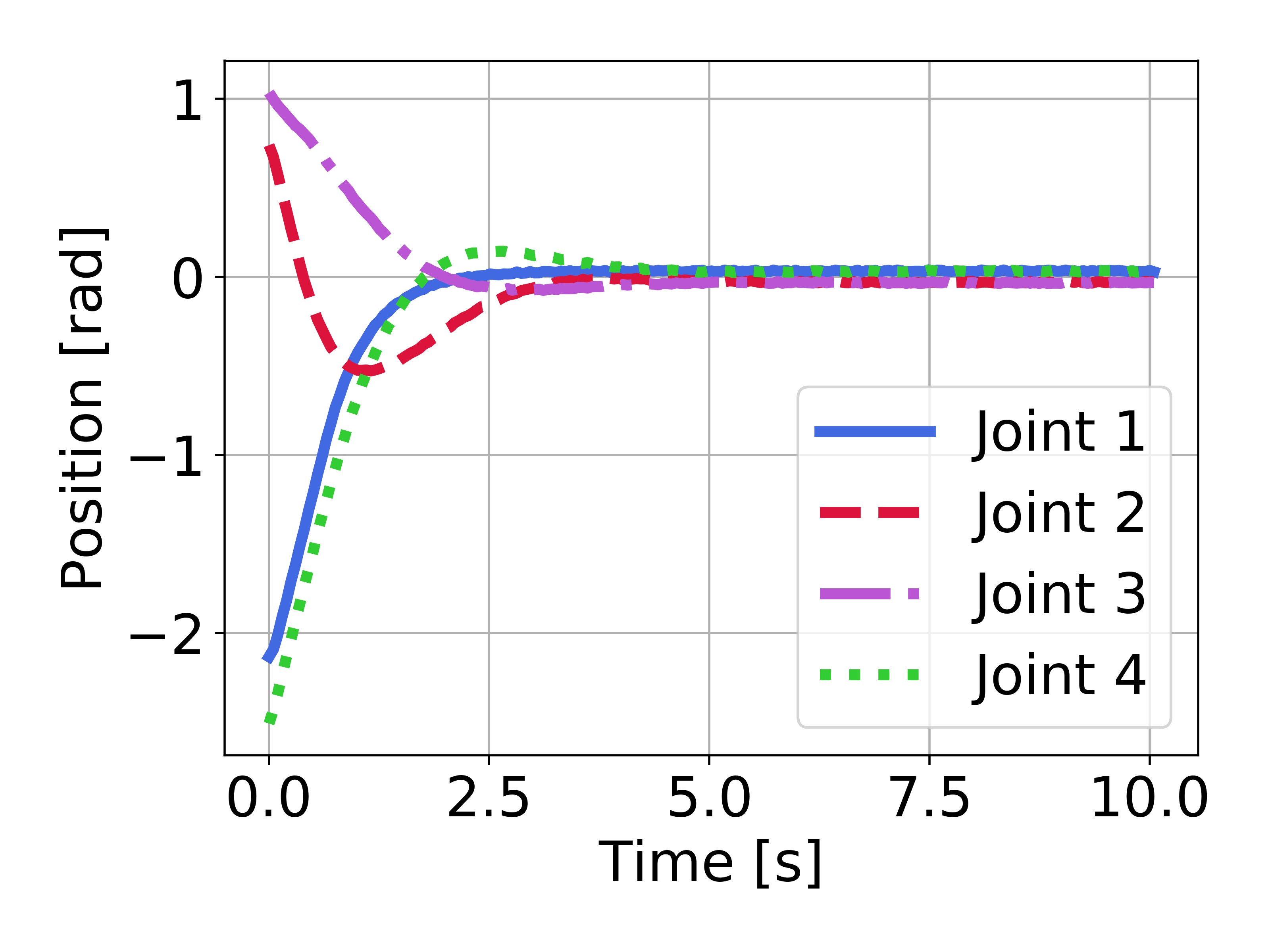}%
        \label{fig:result_compare_mpc_err}%
          } \hfil
    \caption{Comparative results using MPC  for $\mathbf{x}_{ref} = (2.13, -0.74, -1.03,  2.51)$ [rad]}
    \label{fig:result_compare_mpc}
\end{figure*}

\subsection{Comparative results}

In this section we introduce a comparison between the proposed RL agent and a MPC controller. The cost function of the MPC controller is $ J  = \sum_{k = 0}^{N} || \mathbf{x}_{ref} - \mathbf{x}_{t+k} ||^2_{{\textbf{Q}}(t)} + || \Delta \mathbf{u}_{t+ +k}||^2_{{\textbf{R}}(t)}$. The gains $Q$ and $R$ of the cost function where tuned accordingly. As previously, the Reach 5 Alpha simulated arm is used where a desired position ($\mathbf{x}_{ref}$) should be attained.

An experiment is presented when a desired reference position of $\mathbf{x}_{ref} = (2.13, -0.74, -1.03,  2.51)$ [rad] is selected. Fig. \ref{fig:result_compare_rl} shows the obtained joint position when using the proposed RL agent, while Fig. \ref{fig:result_compare_mpc} shows the results when utilizing the baseline MPC controller. While the MPC controller is able to reach the required position for Joint 1 in less than 2.5 seconds (Fig.  \ref{fig:result_compare_mpc_pos}), it takes the rest of the joints longer. On the other hand, the RL agent is faster and presents no overshoot (Fig. \ref{fig:result_compare_rl_pos}). In addition, the RL agent presents no steady state error, as can be seen in Fig. \ref{fig:result_compare_rl_err}, while the MPC shows some error is present in the steady state as Fig. \ref{fig:result_compare_mpc_err} shows. While the control actions both utilize high levels of torque initially, the MPC shown in Fig. \ref{fig:result_compare_mpc_act} seems to require less torque once the steady state is reached as compared to the RL agent, in Fig. \ref{fig:result_compare_rl_act}. 

A series of experiments were performed in which random positions were selected and a number of metrics were obtained in order to compare the performance of the two algorithms. These metrics include the average energy~consumed~(E) in Joules, the Root Mean Square Error (RMSE), Mean Integral Error (MIE), the Mean Steady Steate Error (MSSE), the Overshoot (OS) in percentage, and the Settling Time (ST) in seconds. A set of 20 experiments were conducted where the obtained results can be seen in Table I. 

\begin{table}[!ht] 
\setlength{\tabcolsep}{6pt} 
\renewcommand{\arraystretch}{1.5} 
\centering
\caption{Comparative RL vs MPC}
\begin{tabular}{c | ccccccc}
\hline
Algorithm & E [J] & RMSE & MIE & MSSE &  OS[$\%$] & ST [s] \\ \hline
RL & 19.86 & 0.23 & 58.27 & 0.0033 & 1.43 & 6.26 \\
MPC   & 13.8 & 0.27 & 97.64 & 0.033 & 18.21 & 7.96 \\ \hline
\end{tabular}
\label{tb:result}
\end{table}

The presented metrics show a much more favorable performance of the RL agent. Both MIE and MSSE are significantly lower, while the RMSE also presents lower values for the RL implementation. The OS is practically non existent in the RL agent as compared to the MPC, while the ST is over a second less for the RL, making it the faster solution. The only disadvantage is seen with regards to the energy consumption (E) which is lower for the MPC controller. However, the RL controller is not taking into account the energy consumption as this was not the main focus of the proposal. Overall, the presented results show the benefits of the RL controller. 

\section{Conclusions}
\label{sec:conclusions}
In this article we presented a novel strategy for the low-level control of an underwater manipulator under position and torque constraints based on a reinforcement learning formulation. The actions selected by the actor are directly the torque commands sent to the manipulator, while the state is determined by the current position and velocity, together with the desired joint positions. By including the goal in the state we are able to generalize over different control requests, a fundamental requirement for a control system. 

The data driven approach provided by RL avoids the use of complex models and is able to adapt to changes in the operative conditions.For instance, sudden changes that limit the normal operation of the system, such as obstacles in the working space, failure of any engine, and others, can cause reduced joint movement leading to limited range for joint positions and/or limited torque range. Such constraints can be difficult to surpass using classical controllers due to the lack of accurate model information and poor tuning of the controller. On the other hand, reinforcement learning controllers, are able to obtain highly non linear policies that can operate within the required boundaries and are able to adapt to new requirements. 

As future works, the authors suggest the implementation of the algorithm in the Reach 5 Alpha arm as well as in other manipulators to test the adaptability of the proposal. Furthermore, investigating the possibility of reducing the higher energy consumption, when comparing with the MPC, could be of interest for autonomous operations.

%


\bibliographystyle{unsrt}
{\footnotesize \bibliography{references}}

\end{document}